\documentclass[journal]{IEEEtran}
\usepackage{graphicx}
\usepackage{amsmath}
\usepackage{amsfonts}
\usepackage{algorithm2e}
\usepackage[noend]{algpseudocode}

\makeatletter
\def\BState{\State\hskip-\ALG@thistlm}
\makeatother

\newcommand{\etal}{\textit{et al}.}
\newcommand{\ie}{\textit{i}.\textit{e}.}
\newcommand{\eg}{\textit{e}.\textit{g}.}
\newcommand{\imc}{r}

\ifCLASSINFOpdf
\else
\fi
\begin{document}
%
\title{Grounded and Controllable Image Completion by Incorporating Lexical Semantics}
%
%
%

\author{Shengyu~Zhang$^{*}$,~Tan~Jiang$^{*}$,~Qinghao~Huang,~Ziqi~Tan,~Zhou~Zhao,\\~Siliang~Tang,~Jin~Yu,~Hongxia~Yang,~Yi~Yang,~Fei~Wu
\thanks{S. Zhang, T. Jiang, Z. Tan, Z. Zhao, S. Tang, F. Wu are with College of Computer
Science and Technology, Zhejiang University, Hangzhou, 310027, China (e-mail: sy\_zhang@zju.edu.cn, jiangtan@zju.edu.cn, tanziqi@zju.edu.cn, zhaozhou@zju.edu.cn, siliang@zju.edu.cn, wufei@zju.edu.cn).}
\thanks{Q. Huang, J. Yu, H. Yang are with Alibaba Group, China (e-mail: wfnuser@alumni.sjtu.edu.cn, kola.yu@alibaba-inc.com, yang.yhx@alibaba-inc.com).}
\thanks{Y. Yang is with Faculty of Engineering and Information Technology, University of Technology Sydney, Sydney, Australia (email: Yi.Yang@uts.edu.au)}}

%
%

\markboth{}%
{Shell \MakeLowercase{\textit{et al.}}: Bare Demo of IEEEtran.cls for IEEE Journals}
%



\maketitle

\begin{abstract}
In this paper, we present an approach, namely Lexical Semantic Image Completion (LSIC), that may have potential applications in art, design, and heritage conservation, among several others. Existing image completion procedure is highly subjective by considering only visual context, which may trigger unpredictable results which are plausible but not faithful to a grounded knowledge. To permit both grounded and controllable completion process, we advocate generating results faithful to both visual and lexical semantic context, \ie, the description of leaving holes or blank regions in the image (\eg, hole description). One major challenge for LSIC comes from modeling and aligning the structure of visual-semantic context and translating across different modalities. We term this process as structure completion, which is realized by multi-grained reasoning blocks in our model. Another challenge relates to the unimodal biases, which occurs when the model generates plausible results without using the textual description. This can be true since the annotated captions for an image are often semantically equivalent in existing datasets, and thus there is only one paired text for a masked image in training. We devise an unsupervised unpaired-creation learning path besides the over-explored paired-reconstruction path, as well as a multi-stage training strategy to mitigate the insufficiency of labeled data. We conduct extensive quantitative and qualitative experiments as well as ablation studies, which reveal the efficacy of our proposed LSIC.
\end{abstract}

\begin{IEEEkeywords}
image Completion, adversarial learning, text guided, controllable generation
\end{IEEEkeywords}

%
\IEEEpeerreviewmaketitle

\renewcommand{\thefootnote}{\fnsymbol{footnote}}
\footnotetext[1]{These authors contributed equally to this work.}
\renewcommand{\thefootnote}{\arabic{footnote}}

\section{Introduction}

\begin{figure}[!t]
\centering
\includegraphics[width=\columnwidth]{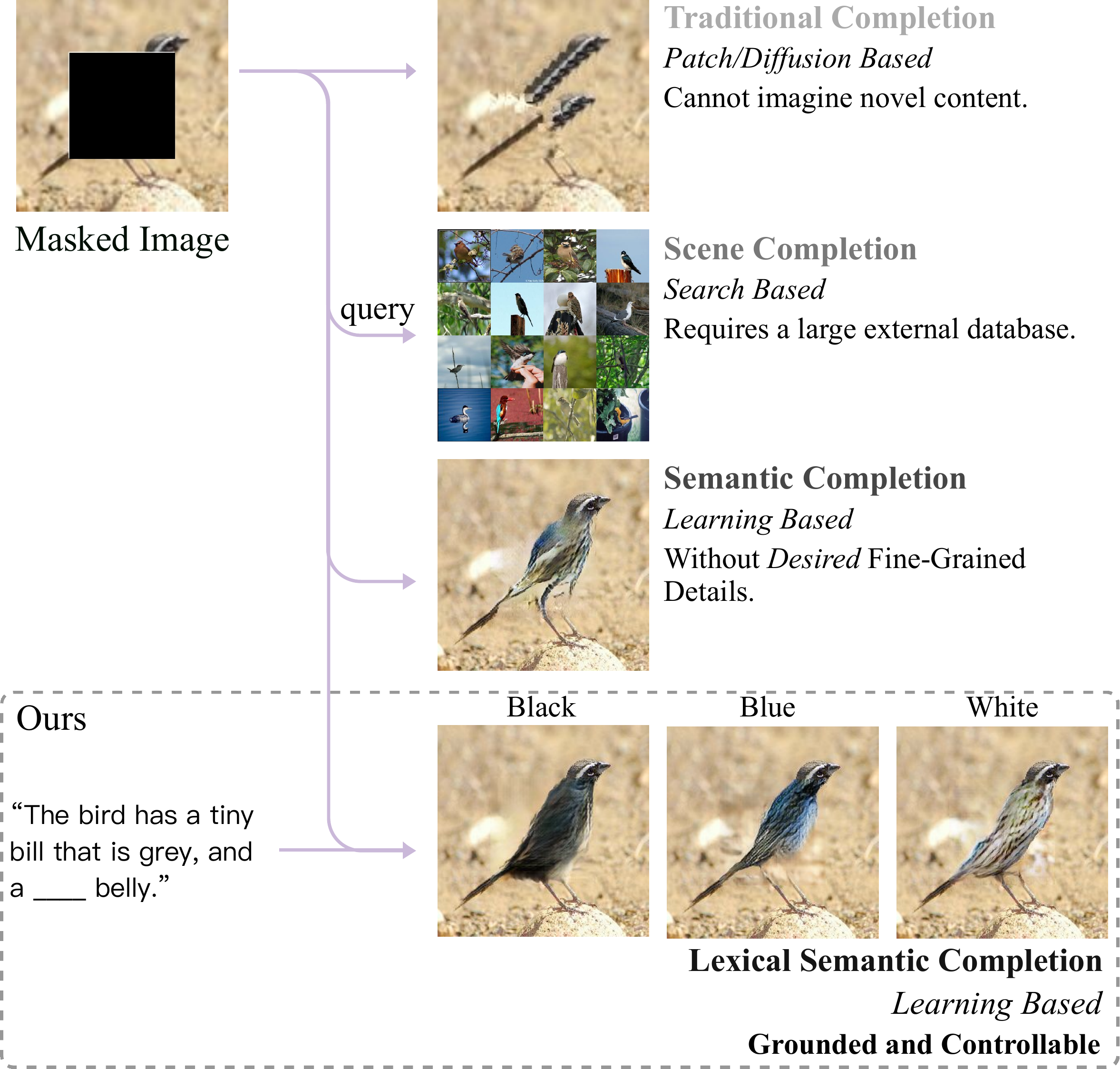}
\caption{
Overviews of Lexical Semantic Image Completion and previous completion methods. 
}
\label{fig:fristpage}
\end{figure}

\IEEEPARstart{A}{mong} numerous research fields of computer vision, image completion is a longstanding and fundamental problem and has become ubiquitous in diverse real-world applications, such as restoration of damaged paintings and object removal. Bertalmio \etal \cite{bertalmio2000image} summarized the restoration process as four sequential steps: 1) Imagine according to the global picture; 2) Ensure the continuity of the gap between the masked and unmasked area; 3) Depict contour lines and colors; 4) Add some details like texture and patterns. Traditional completion techniques \cite{bertalmio2000image,levin2003learning,barnes2009patchmatch,Liu_Long_Zhu_2019,Liu_Yang_Fang_Guo_2018} attempt to match similar patches and copy them to masked regions. These kinds of methods can ensure Step 2, but they may fail to obtain a reasonable imagination at Step 1 or depict meaningful structures at Step 3 due to the absence of the high-level understanding \cite{zeng2019learning}. Scene completion methods \cite{hays2007scene,zhu2015faithful,zhao2019unconstrained} require a large-scale image database covering possible completion targets, which is not always available. The recent progress in deep neural networks has shown high capability in learning generative models to accomplish the main part of the above inpainting process \cite{pathak2016context,iizuka2017globally,yang2017high,yeh2017semantic,zeng2019learning,sagong2019pepsi}. However, on the one hand, these completion techniques tend to generate blurry regions and artifacts \cite{yan2018shift,song2018contextual,liu2018image,yu2018generative}, especially when the hole is rather large, due to lack of information of foreground objects\cite{xiong2019foreground}.

On the other hand, the subjective nature \cite{bertalmio2000image,Zheng:2019tt} of image completion may lead to results that are visually authentic but not faithful to a grounded truth (\ie, factual cues or attribute information). The \textit{grounded} and \textit{controllable} image completion can be a fundamental requirement in many real-world scenarios. For example, in the conservation of historic and artist works, automatic tools should be developed to understand some old written pieces, \eg Chinese poems to an ancient painting, that contain the detailed information of the damaged paintings. For object replacement in image design and editing, users may usually require an iterative and controllable completion process, and natural language offers a general and flexible interface to describe her/his intentions.

To bridge these gaps, we propose an approach named \textit{Lexical Semantic Image Completion (LSIC)}. The completion results are conditioned not only on the structural continuity and visual semantic but also on the lexical semantic concepts within natural language descriptions. In such a setting, we aim to obtain a grounded imagination at Step 1 and a more controllable process in Step 3 and 4. Fig. \ref{fig:fristpage} illustrates a comparison between our method and previous methods, i.e., traditional completion method \cite{criminisi2004region}, scene completion and semantic completion \cite{Zheng:2019tt}.


\begin{figure}[!t] \begin{center}
    \includegraphics[width=\columnwidth]{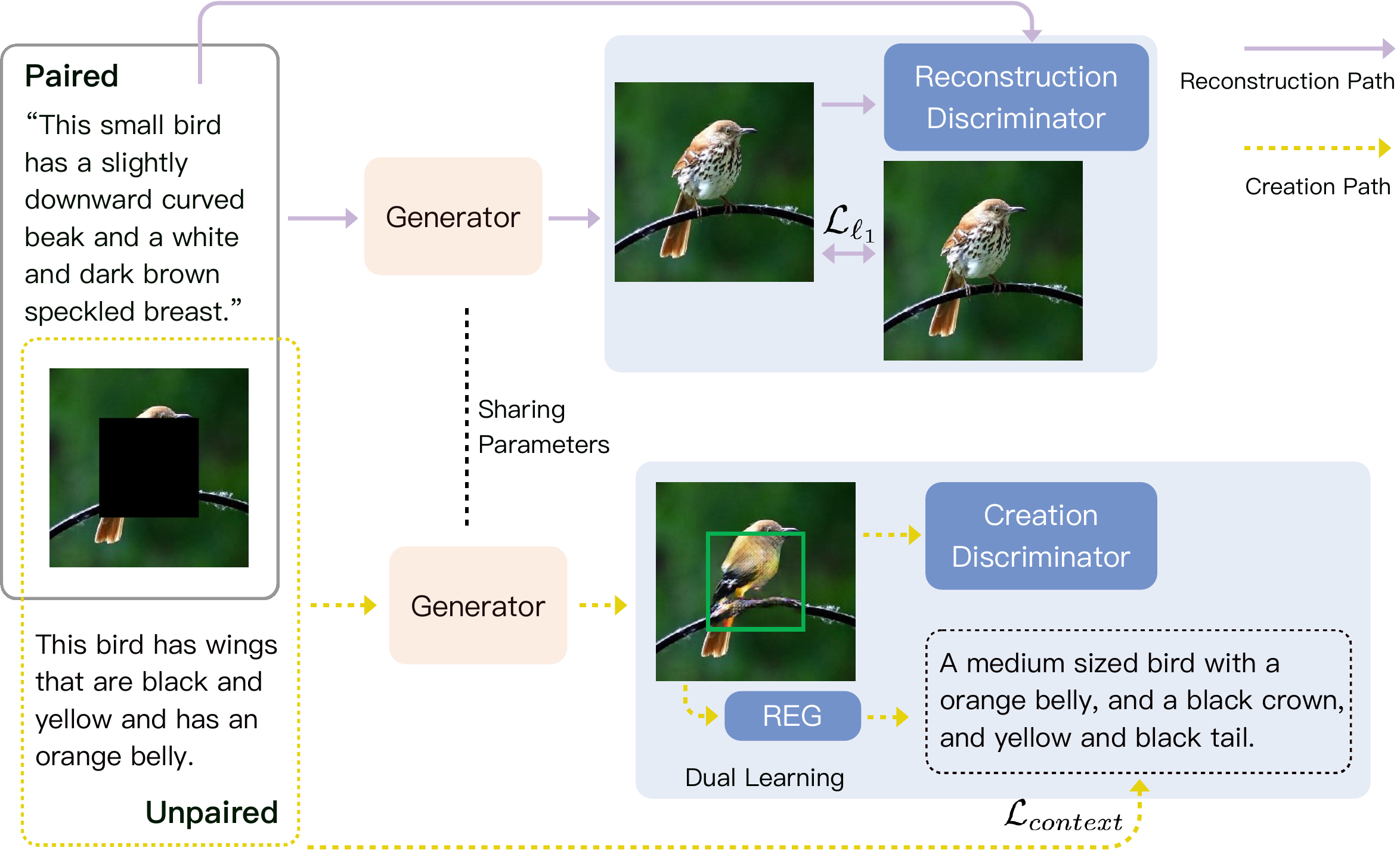}
    \caption{
    Overall schema of our model, which mainly comprises a progressive reasoning generator and two learning paths.
	}
\label{fig:schema}
\end{center} \end{figure}




One major challenge of LSIC is the sheer difficulty to model both the visual semantic structure within the unmasked image and the lexical semantic structure within the sentence and to learn the aligned relationship between them. The visual semantic structure within the masked region can be reasoned and inferred from the aligned relationship eventually. We denote this process as \textit{structure completion}. To address this challenge, we propose to harness the flexible Graph Convolution Network \cite{kipf2016semi} to perform structure message passing and construct several reasoning blocks. Moreover, by investigating the empirical completion process, especially Step 3 (draw globally) and Step 4 (draw locally), we propose first to perform coarse-grained reasoning to depict rough shapes and colors and refine it \textit{progressively} by performing fine-grained reasoning, which is realized by coarse-grained reasoning block (CGR) and fine-grained reasoning block (FGR) in our model. This scheme avoids the complexity of capturing the high-level concepts and elaborate details in one place by divide-and-conquer.

Another challenge regards collecting dataset containing multiple text conditions per masked image, which is often prohibitively expensive to acquire or even unavailable. The annotated sentences for one image are often semantically equivalent in existing datasets. The only-one-text may lead to \textit{unimodal biases}, which is common in the VQA task\cite{cadene2019rubi}. In other words, the completion result may be mostly conditioned on the unmasked regions while disregarding the text information, resulting in the loss of controllable generation ability or suffering a performance drop on the test dataset. To this end, we consolidate the idea of Dual Learning \cite{he2016dual} and devise an \textit{unpaired-creation} training path besides the conventional \textit{paired-reconstruction} training path. The unpaired-creation path takes a masked image and a randomly sampled text as input. Since there are no labeled completion results, we incorporate a Referring Expression Generation \cite{kazemzadeh2014referitgame} method to re-generate the input texts, which can be considered as the dual task of ours. The re-generation error provides informative feedback to guide this unpaired data learning. Based on this training schema, we aim to enable controllable image completion and improve generalization ability.


To summarize, this work makes the following key contributions:

\begin{itemize}
	\item We propose a new approach termed \textit{Lexical Semantic Image Completion}, which aims to enable \textit{grounded} and \textit{controllable} completion.
	\item We devise two novel reasoning blocks, \ie CGR and FGR, to perform \textit{structure completion} progressively.
	\item We formulate an \textit{unpaired-creation} learning path as a counterpart to the common paired-reconstruction path to reduce \textit{unimodal biases} and improve generalization.
\end{itemize}











%

%
%
%

\section{Related Works}

%

\paragraph{Semantic Image Completion}


First proposed by Pathak \etal\cite{pathak2016context}, Semantic Image Completion (SIC) \cite{pathak2016context,yeh2017semantic,li2017generative,liu2018image,yu2018free,yang2017high,yu2018generative,yan2018shift,song2018contextual} exploits information from the whole dataset besides the unmasked. There are works \cite{Lie_Hsieh_Lin_2018,Li_Zhu_Sun_2018,Nguyen_Kim_Hong_2019,ller_Wiegand_2011,Macchiavello_Dorea_Hung_Cheung_Tan_2014} exploiting extra reference images, \eg, other frames in the same video, or additional depth information as the semantic context. The Semantic in existing works refers to the image semantics or high-level context in the dataset. In contrast, the Semantic in our task refers to the lexical semantics, \ie, the concepts and meanings within natural language.  LSIC provides a natural interface to incorporate detailed requirements and necessary knowledge in the form of natural language into the completion procedure, aiming to enable grounded and controllable completion.
The idea of progressive image completion has been explored in previous approaches \cite{DBLP:conf/mm/ZhangHLZW18,DBLP:conf/mm/Guo0YCL19}. They complete the hole by progressively reducing the masked area. Differently, we first capture the global picture and focus on fine-grained details subsequently to progressively learn a comprehensive representation of different modalities. We are driven by different motivations, and our architecture is designed explicitly for LSIC.

\begin{figure*}[t] \begin{center}
    \includegraphics[width=\textwidth]{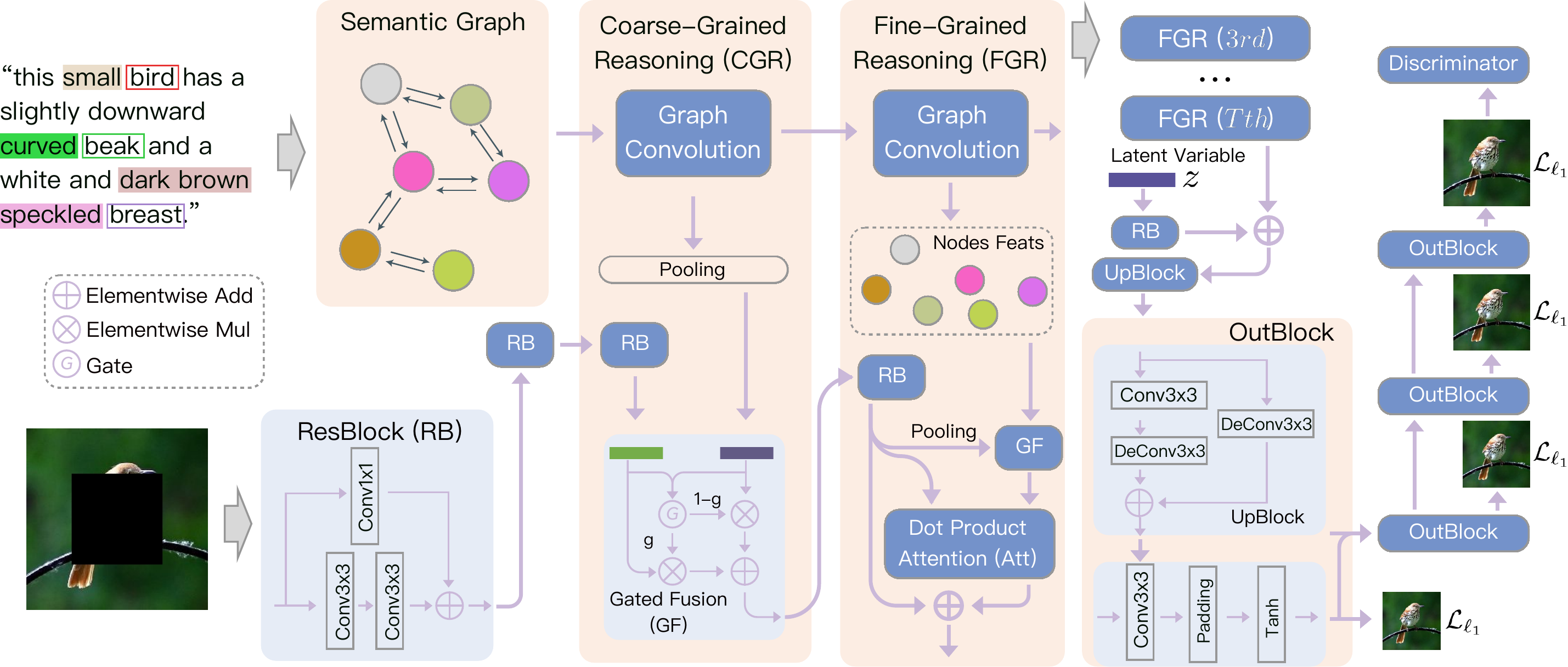}
    \caption{
    The detailed architecture of the generator, which comprises one coarse-grained reasoning block and multiple hierarchically stacked fine-grained reasoning blocks. 
	}
\label{fig:generator}
\end{center} \end{figure*}

\paragraph{Semantic Image Synthesis}

In the research area of image generation/synthesis, semantic image manipulation (SIM) is a similar task to ours, which refers to image manipulation according to natural language descriptions. Dong \etal \cite{DBLP:conf/iccv/DongYWG17} firstly proposed an end-to-end neural architecture built upon the GAN \cite{goodfellow2014generative} framework to change the color of objects semantically. TAGAN \cite{nam2018text} devises a text-adaptive discriminator to better preserve text-irrelevant contents. MC-GAN \cite{DBLP:conf/bmvc/ParkYK18} introduced a Multi-conditional GAN to semantically generate an object on a complete base image with the help of object segmentation annotation. The main difference between our method and previous arts is that LSIC requires the model to be able to generate the background, the whole object, the partial object, or just a fine-grained detail since the mask can vary while SIM often has a well-defined goal as defined in the above works, such as color change or adding a foreground object to a complete background. 



%
%
%
%
%

\section{Method}


In this section, we will elaborate on each building block of our proposed model. As shown in Fig. \ref{fig:schema}, our proposed schema mainly embodies a \textit{coarse-to-fine generator}, a \textit{reconstruction path}, a \textit{creation path} and a \textit{referring expression generator} (REG). We begin with the overall schema and some notations. Let $I$, $I_m$, and $\hat I$ denote the original image, the partially masked image, and the generated image, respectively. $t$, $ \bar t $ and $\hat t$ denote the paired text to $I$, an unpaired text which is randomly sampled and the re-generated text by REG, separately. Our model encapsulates a conventional reconstruction path, which takes $I_m$ and $t$ as inputs and uses $I$ as the ground truth in a supervised manner. As a counterpart, the unsupervised creation path takes $I_m$ and $\bar t$ as inputs. Both paths are trained adversarially while the reconstruction discriminator is a conditional one, and creation discriminator is not. The creation path incorporates a referring expression generator to guarantee the faithfulness of the generation to the text condition. The generators in two paths have the same structure and share weights. Notations are summarized in Table \ref{tab:notations}.


\subsection{Generator}

Fig. \ref{fig:generator} shows the detailed architecture of the generator. The generator takes text $t$ or $ \bar t $ and the masked image $I_m$ as inputs, aiming to generate a realistic-looking, globally coherent, and locally consistent image $ \hat I $ grounded in the text condition. We propose to explicitly model the lexical semantic structure for \textit{Structure Completion}. Specifically, we use StanfordNLP toolkit \cite{qi2018universal} to extract dependency relation between words in description sentences and transform text $t$ or $\bar t$ into graph representation $g$ or $\bar g$. Since the semantic relation between nodes cannot be simply evaluated by the similarity of word embedding vectors, we initialize all the edge weights by default 1. The masked image is encoded by two standard resnet blocks into an image code $\imc_0 = \{ \imc_i  \}_{i=1, \ldots, N_{0,r}}$, where $N_{0,r}$ is the number of initial visual region features. 

To perform structure alignment and translation both globally and locally, we devise two novel reasoning blocks, termed coarse-grained reasoning block (CGR) and fine-grained reasoning block (FGR). This schema is designed as progressive and multi-grained. 

\begin{table}[!t]
\centering

\setlength\doublerulesep{0.5pt}
\small
\caption{
    Notations
}
\begin{tabular}{c|l}
\hline
\hline
\multicolumn{1}{c|}{ \textbf{Notation} }        & \multicolumn{1}{l}{ \textbf{Description} }          \\
\hline \hline
$I,I_m,\hat I$     & the real/masked/generated image \\

$t,\bar t,\hat t$         & the input/randomly-sampled/re-generated text          
\\
$g,\bar g$   &  the semantic graph built upon $t,\bar t$        \\

$V,v$ & the set of node features, the feature of one node \\

$r,c$ & the intermediate visual features \\

$v_{*}, c_{*}$  &  the mean-pooled nodes/visual features $$    \\

$N_v,N_r$    & the number of nodes/visual features \\

$T, \tau$ & the number of reasoning blocks, the current step \\

\hline
\end{tabular}
\label{tab:notations}
\end{table}


\paragraph{Coarse-Grained Reasoning block} Starting from the initial image code $\imc_0$ and the initial semantic graph $g_0 = (V_0, E_0)$. The CGR firstly performs visual structure reasoning by the resnet block and obtains $c_1 = \{ c_i \}_{i=1, \ldots, N_{1,r}} $. For lexical semantic structure modeling, CGR employs the Graph Convolution Network (GCN) to reason along the grammar connection between words and thus generates features $V_1 = \{ v_{1,j} \}_{j=1, \ldots, N_v}$ with the semantic relationship, where $N_v$ is the number of nodes or words. Then, as a common practice, we obtain the high-level semantic concepts by pooling the graph into a global representation $v_{1,*}$. This graph representation can help draw coarse-grained shapes, such as the body structure, and infer the main color palette. This process can be formulated as:
\begin{align}
	V_1 &= \mathbf{D}^{-1/2} \mathbf{A}
        \mathbf{D}^{-1/2} V_0 \mathbf{\Theta}, \\
	v_{1,*} &= Pool(V_{1}) =1/N_v \sum_j v_{1,j}
\end{align}
where $\mathbf{\Theta}$ is the weight of graph convolutions and $\mathbf{A}$ is the adjacency matrix with inserted self-connections. $\mathbf{D}$ denotes the diagonal degree matrix for $\mathbf{A}$ and is computed as $\mathbf{D}_{uu} = \sum_{v=0} \mathbf{A} _{uv}$. And thus $\mathbf{D}^{-1/2} \mathbf{A} \mathbf{D}^{-1/2}$ denotes the standard row-wise normalization. The $Pool$ function used in our model is mean-pooling.

LSIC, at a basic level, is to extract relevant visual-semantic information to fill the hole, such as style, patterns, texture, lines in the image, and descriptive details in words. In the CGR block, this sense of filtering relevant content is realized using a \textit{gated fusion} function. Given the image region feature $\imc_{1,i}$ and graph representation $v_{1,*}$, the gated fusion function performs the following operations:
\begin{align}
	\alpha_i &= \sigma(W_{1,a}[c_{1,i}, v_{1,*}]) \\
	\imc_{1,i} &= \alpha_i * W_{1,\imc}c_{1,i} + (1-\alpha_i) * W_{1,g}v_{1,*}
\end{align}
where $\sigma$ is the sigmoid function. $[.,.]$ denotes the concatenate operation. $W_{1,a}$ and $W_{1,r}$ and $W_{1,g}$ are linear transformations. $W_{1,r}$ and $W_{1,g}$ transform image features and  graph features into a joint visual-semantic feature space. Intuitively, this function will activate graph-level relevant regions (bigger gate score $\alpha_i$) and suppress the irrelevant ones (smaller gate score) through pair-wise end-to-end training., which helps extract useful visual-semantic information such as style and patterns.

\paragraph{Fine-Grained Reasoning block}
The aforementioned coarse-grained reasoning block considers high-level semantics and structural continuity like contour lines and colors while neglecting the fine-grained detail like texture and patterns. To this end, we design the fine-grained reasoning block to capture these details using node-level attention. In the $\tau {th}$ iteration, FGR takes the nodes features $V_{\tau-1}$ and image features $\imc_{\tau-1}$ from previous ${(\tau-1)} {th}$ reasoning block as input. Similarly, FGR performs $c_{\tau} = ResBlock(\imc_{\tau-1})$ and $V_{\tau} = GCN(V_{\tau-1}, E_{\tau-1})$ for visual-semantic structure modeling. Different from CGR, FGR builds the gated fusion function as follows:
\begin{align}
	&c_{\tau,*} = 1/N_{\tau,r} \sum_i c_{\tau,i} \\
	&\beta_{\tau,j} \ = \sigma(W_{\tau,a}[c_{\tau,*}, v_{\tau,j} ]) \\
	&o_{\tau,j} = \beta_{\tau,j} * W_{\tau,r} c_{\tau,*} + (1-\beta_{\tau,j}) * W_{\tau,g}v_{\tau,j}
\end{align}
While the gated fusion function in CGR is designed to filter useful patterns in the unmasked visual context, the gated fusion function in FGR is specially designed to filter the fine-grained semantic concept ready for further aggregation. Given the fused features $o_{\tau} = \{ o_{\tau,j} \}_{j=1, \ldots, N_v}$ and image features $c_{\tau}$, we apply attention mechanism to perform local visual-semantic reasoning on salient and reusable visual patterns as well as meaningful semantic concepts, and thus aligning the structures of two modalities. It also helps suppress unnecessary background regions and irrelevant lexical information. For $i_{th}$ image region feature, we compute the lexical-semantic-aware visual features as:
\begin{align}
	\epsilon_{\tau,i,j} &= \frac{ \exp(f(c_{\tau,i}, o_{\tau,j})) }{\sum_k \exp(f(c_{\tau,i}, o_{\tau,k})) }
\end{align}
\begin{align}
	r_{\tau,i} &= c_{\tau,i} +  \sum_{j=1}^{N_v} \epsilon_{\tau,i,j} W_{\tau,m} o_{\tau,j}
\end{align}
where function $f$ computes the joint-space similarity of $c_{\tau,i}$ and $o_{\tau,j}$ by $f(c_{\tau,i}, o_{\tau,j}) = (W_{\tau,l} c_{\tau,i})^T (W_{\tau,n} o_{\tau,j}) $. $W_{\tau,m}$, $W_{\tau,l}$ and $W_{\tau,n}$ are linear transformations at step $\tau$. Compared with CGR, FGR further incorporates word-level attention to score image region features and weighted-sum them, which captures the unit-level details in both visual and lexical semantics. 
The first CGR and following $T-1$ FGRs are stacked sequentially and form the multi-grained progressive generation process. The randomly sampled latent variable $z$ is added after the stacked reasoning blocks. Our generator incorporates multiple output layers to generate multi-scale images hierarchically, as depicted in Fig. \ref{fig:generator}.


\subsection{Discriminator and Two Learning Paths}

As shown in Fig. \ref{fig:schema}, our framework comprises two parallel training paths, \ie, the supervised \textit{reconstruction} path, and the unsupervised \textit{creation} path.

\paragraph{Reconstruction Path}  The reconstruction path follows the conventional pipeline, which takes the masked image $I_m$ and the paired textual description $t$ as input. We train it adversarially using a conditional discriminator $D_R$, which grades not only the visual plausibility but also whether the image is well conditioned on the text constraint. We name it reconstruction path since there is a paired ground-truth image, and we add hierarchical $\ell_1$ losses in different scales. During training, the loss function introduced by this path can be defined as:
\begin{align}
	\mathcal{L}_{G}^R = &\underbrace{ - \lambda_{adv} \mathbb{E}_{\hat I \sim p_{G}} \log{ D_R(\hat I, v_{0,*}) } }_{ \text{ conditional adversarial loss } } + \underbrace{ \lambda_{l_1} \mathbb{E}_{\hat I \sim p_{G}} || I - \hat I ||_1 }_{ \ell_1 \text{ loss } } \\
	{\mathcal{L}_{D}^R} = 
	&- \mathbb{E}_{ I \sim p_{data} } \log{ D_R(I, v_{0,*}) }
	 - \mathbb{E}_{ \hat I \sim p_{G} } \log{ (1 - D_R( \hat I, v_{0,*})) }
\end{align}
where $v_{0,*}$ denotes the initial global representation of semantic graph, which is obtained by $v_{0,*} = Pool(V_0)$. The pooling method used in our paper is mean pooling. We omit the multi-scale $\ell_1$ losses for brevity.

\paragraph{Creation Path} The Lexical Semantic Image Completion task naturally requires the ability to generate different images given different text constraints with the same background, \ie, controllable completion. However, the number of paired sentences to one masked image is typically only one, or they are semantically equivalent. The only one ground-truth also underestimates the variety of results. 

To this end, by consolidating the idea of leveraging dual learning in unpaired sequence-to-sequence translation, we propose an unsupervised creation path. The creation path takes the unpaired textual description and masked image as input. Since there is no ground-truth image, we employ an unconditional discriminator to guarantee the visual plausibility. We did not employ a conditional discriminator because it may be hard for the discriminator to simultaneously capture both visual quality and faithfulness to textual constraint at one place with unpaired input. We incorporate a referring expression generator to re-generate the description with the unmasked area as context. The cross-entropy loss between the re-generated words and the input tokens penalizes the inconsistency between the completion area and semantic-visual context, \ie the input text and unmasked area. Therefore, we name it context loss. The loss function introduced by creation path can be formulated as:
\begin{align}
		\mathcal{L}_{G}^C = & \underbrace{ - \lambda_{adv} \mathbb{E}_{\hat I \sim p_{G}} \log{ D_C(\hat I) } }_{ \text{ adversarial loss } } \underbrace{ - \lambda_{ce} \mathbb{E}_{\hat I \sim p_{G}} \sum\nolimits_{ \kappa } \log P( \hat t_{ \kappa } ) }_{ \text{ context loss } } \\
			{\mathcal{L}_{D}^C} = 
	&- \mathbb{E}_{ I \sim p_{data} } \log{ D_C(I) } - \mathbb{E}_{ \hat I \sim p_{G} } \log{ (1 - D_C( \hat I)) }
\end{align}
where $\hat t = \{ \hat t_{\kappa} \}_{{\kappa}=1, \ldots, N_e} = REG(\hat I) $ is the re-generated sentence and $N_e$ is the length of the sentence. Two typical works that attempt to incorporate the dual learning idea into image generation are CycleGAN \cite{zhu2017unpaired} and MirrorGAN \cite{qiao2019mirrorgan}. Despite the similar structure, we are driven by different motivations, and there are several crucial differences: 1) Our model incorporates both supervised and unsupervised learning path while CycleGAN is trained merely unsupervisedly and MirrorGAN is trained totally supervisedly; 2) Our schema is not \textit{cycle structure} or \textit{mirror structure} since we only re-generate the partial input (the text) with another input (the unmasked image) as a visual context using the REG. Our design is explicitly designed to accommodate the Unimodal Biases problem for LSIC.


\paragraph{Discriminator Architecture} We follow a standard Patch GAN architecture \cite{Isola_Zhu_Zhou_Efros_2017} to build the discriminators in two paths. The building blocks are of the same structure as the $ResBlock$ in the generator. This design is fully convolutional, which is compatible with arbitrary-scale input image and thus compatible with arbitrary number of FGR blocks. The conditional discriminator in the paired-reconstruction path considers the text conditions by concatenating the intermediate downsampled image feature map with the graph-level representation $v_{0,*}$, following \cite{zhang2017stackgan}.  

\paragraph{Referring Expression Generation} We propose a simple architecture for REG, which builds on the image caption architecture \cite{xu2015show}. Given the generated image $I$ and a bounding box indicating the target area, we extract features from $I$ as context code using CNNs. The target area is up-sampled to the same size as $I$ and encoded as the target code using the identical CNNs. We concatenate the target code with the context code and send them into the decoder (RNN), which generates the final referring expression.


\begin{figure*}[t] \begin{center}
    \includegraphics[width=1.0\textwidth]{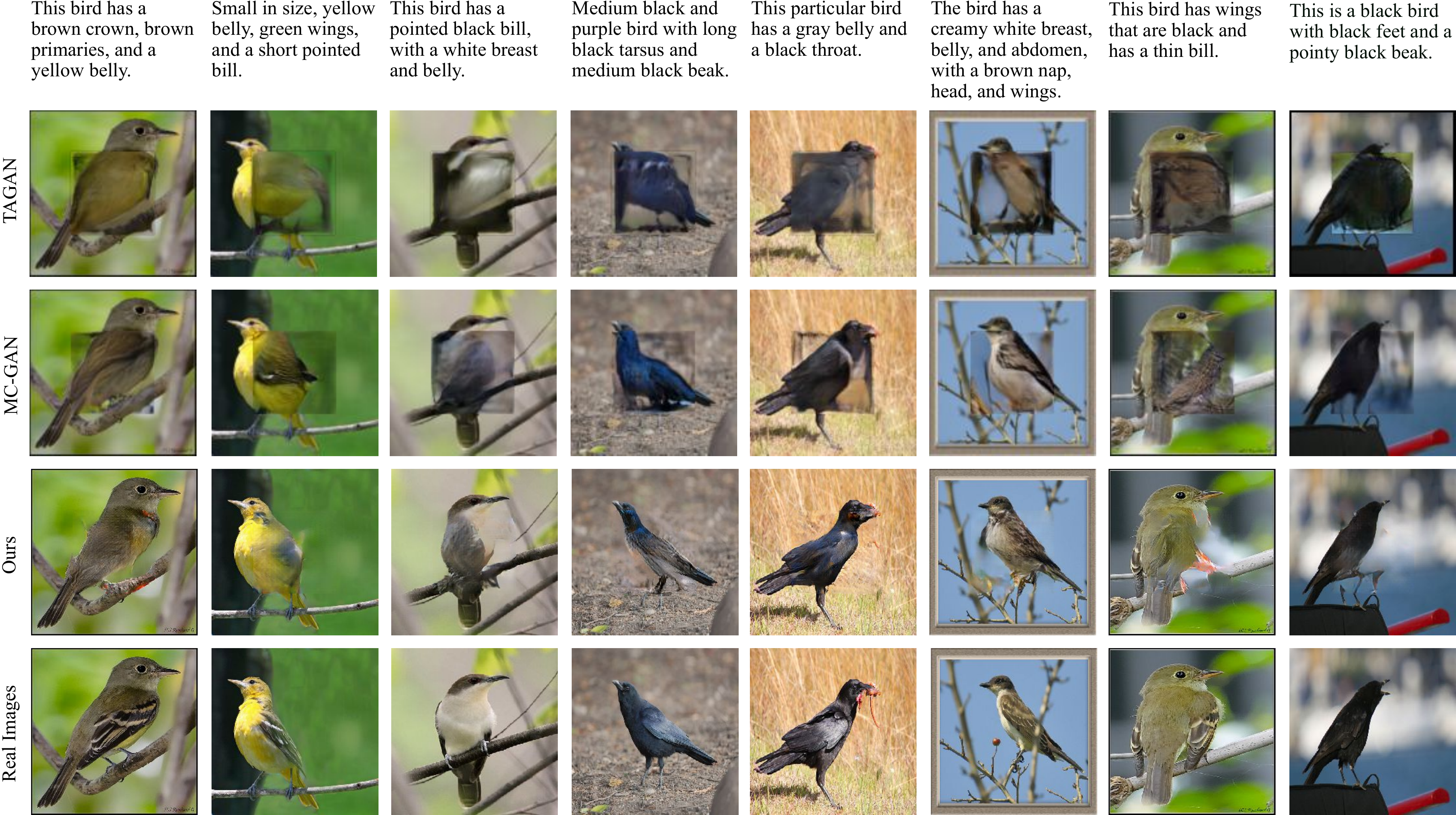}
    \caption{
    Qualitative comparison of three methods on the CUB test set with centered square mask (50\% * 50\%). Our models generate completions with the best overall quality and grounded details.
	}
\label{fig:compare}
\end{center} \end{figure*}

\begin{table*}[t]
\centering
\caption{
     Quantitative results on the CUB test set and Oxford-102 dataset.
}
\setlength{\tabcolsep}{5.5pt}
\setlength\doublerulesep{0.5pt}
\begin{tabular}{l|cccc|cccc|cccc}
& \multicolumn{4}{c}{CUB (Center)} & \multicolumn{4}{c}{ CUB (Free-form)}  & \multicolumn{4}{c}{ Oxford-102 (Center) }   \\


\multicolumn{1}{c|}{Method}     & PSNR $\uparrow$         & TV loss $\downarrow$   & SSIM $\uparrow$          & IS  $\uparrow$       & PSNR $\uparrow$         & TV loss $\downarrow$   & SSIM $\uparrow$          & IS  $\uparrow$      &  PSNR $\uparrow$         & TV loss $\downarrow$   & SSIM $\uparrow$          & IS  $\uparrow$                \\
\hline \hline
Dong  \etal        & 14.63      &    15.08   &    0.70     &    2.71    & 16.75 &  14.19 & 0.75  & 2.96   & 13.89      &    16.04   &    0.71     &    3.07         \\
TAGAN   &  19.10   & 13.42    & 0.76   &    4.04    &  23.96   &      13.05   &  0.83   &   4.11    &  19.50   & 13.92    & 0.78   &    3.89  \\
MC-GAN   & 18.23    & 14.88    &   0.75  &   3.98   &  24.30  &  13.27   &  0.82  &  4.20    & 19.50    & 15.69    &  0.76  &   4.31  \\
\hline \hline
Ours  & \textbf{19.68}    &   \textbf{10.73}   &    \textbf{0.82}   & \textbf{4.34}    &   \textbf{26.19} & \textbf{10.87} & \textbf{0.90} &   \textbf{5.82}   & \textbf{19.83}    &    \textbf{10.99}   &      \textbf{0.81}   & \textbf{5.28}    \\
\hline
\end{tabular}

\label{tab:quantitative}
\vspace{-0.3cm}
\end{table*}
%

\begin{algorithm}[t]
\footnotesize
 \KwData{A set of image-sentence pairs. Epoch number $N^{Sup}$, $N^{Warm}$ and $N^{Har}$ for the \textit{Supervision}, \textit{Warm-start} and \textit{Harmony} stage, individually. Batch number $N^{Batch}$ in one epoch.}
 \KwResult{Optimized generator $G$, reconstruction discriminator $D_R$, creation discriminator $D_C$ and referring expression generation model $REG$.}
 \SetKwInput{KwData}{\textit{Supervision Stage}}
 \KwData{}
 Initialize $G$ and $D_R$ at random.
 
 \For{$s$ = $1:N^{Sup}$}{
   Sample paired data batches. Train $G$ and $D_R$ by minimizing $\mathcal{L}_{G}^{R}$ and $\mathcal{L}_{D}^{R}$, alternatively.
 }
 
Train $REG$ until convergence using the cross-entropy loss.
 
 \SetKwInput{KwData}{\textit{Warm-start Stage}}
 \KwData{}
 
 Initialize $D_C$ at random and freeze the parameters of $G$.
 
 \For{$s$ = $1:N^{War}$}{
 
 	\For{$b = 1:N^{Batch}$}{
   		Sample one batch of unpaired data and train $D_C$ by minimizing $\mathcal{L}_{D}^{C}$.
   		
   		Using the generated images and the corresponding sentences to fine-tune the $REG$.
	}

 }

 \SetKwInput{KwData}{\textit{Harmony Stage}}
 \KwData{}
Unfreeze the parameters of $G$ for training.
 
 \For{$s$ = $1:N^{Har}$}{
   
   \For{$b = 1:N^{Batch}$}{
   		Sample one batch of paired data. Train $D_R$ and $G$ by minimizing $\mathcal{L}_{D}^{R}$ and $\mathcal{L}_{G}^{R}$, alternatively.
   		
   		Sample one batch of unpaired data. Train $D_C$ and $G$ by minimizing $\mathcal{L}_{D}^{C}$ and $\mathcal{L}_{G}^{C}$, alternatively.
   }

 }

 \caption{Multi-stage training procedure.}
 \label{alg:training}
\end{algorithm}

\subsection{Training Procedures}

Since the creation path is entirely unsupervised, it may be hard to produce useful gradients at the very beginning. To this end, we split the training of our model into three stages. (See algorithm \ref{alg:training}) In the first \textit{Supervision} stage, we train the generator and reconstruction discriminator with $\mathcal{L}_{G}^{R}$ and $\mathcal{L}_{D}^{R}$. The REG module is also pre-trained. In the next \textit{Warm-start} stage, we fix the parameters of the generator and reconstruction discriminator and train the creation discriminator with $\mathcal{L}_{D}^{C}$. We also "fine-tune" the REG module with the generated fake images as input and make the input distribution align to the present generation distribution. In the last \textit{Harmony} stage, we co-train all the modules with all losses to further improve the generation quality and increase the diversity. In our experiments, we observed that this strategy leads to more robust training and less fitting iterations. It is worth noting that multiple creation learning paths can be constructed to explore possible background-text combinations simultaneously. Different learning paths, including one reconstruction path and multiple creation paths, can be trained in a distributed and parallel way to obtain both efficiency and effectiveness, which can be a promising future work.

\subsection{Implementation Details}


Our generator contains one CGR block and two stacked FGR blocks. The whole network is trained using Adam optimizer \cite{kingma2014adam} with batch size as $8$, $\beta_1=0.0$ and $\beta_2=0.999$. The learning rate is fixed to $0.0001$. $(\lambda_{adv}, \lambda_{l_1}, \lambda_{ce})$ is set to $(1, 20, 1)$. We use Orthogonal Initialization \cite{saxe2013exact} and apply Spectral Normalization \cite{miyato2018spectral} in all networks. No data augmentation technique applies to our training. We set $N^{Sup}$, $N^{Warm}$ and $N^{Har}$ to 20, 20 and 30, respectively.

\section{Experiments}

\paragraph{Datasets}

We mainly carry out experiments on two fine-grained caption-annotated dataset, CUB \cite{wah2011caltech} and Oxford-102 \cite{nilsback2008automated}. CUB contains 11,788 images over 200 different categories while Oxford-102 has 8,189 images over 102 categories. Following \cite{reed2016learning}, we split these two dataset into category-disjoint training/test sets and use 5 captions per image as in \cite{reed2016generative}. We obtain the word embedding using Glove \cite{pennington2014glove} vectors of version cased-300d.


\paragraph{Compared methods and baselines}
	
Since there is no research precisely comparable to this paper, we adopt three state-of-the-art semantic image manipulation methods and make proper adjustments to them. Specifically, we incorporate Dong \etal \cite{DBLP:conf/iccv/DongYWG17}, MC-GAN \cite{DBLP:conf/bmvc/ParkYK18}, and TAGAN \cite{nam2018text}. Dong \etal and TAGAN concern attribute transfer according to natural language descriptions. The difference between input and output in this task is often subtle compared to our completion task. Therefore, we mainly add $\ell_1$ loss to ease the training and obtain structural continuity. MC-GAN attempts to add a separate object to a complete background according to textual description, which is more close to our task. To get a fair comparison, we remove the segmentation estimator and segmentation encoder in MC-GAN, which requires additional segmentation annotation.

\subsection{Quantitative Evaluation}
	
Following the image completion convention, we choose to evaluate the generation results with three numeric metrics, \ie, Peak Signal-to-Noise Ratio (PSNR), Total Variation (TV) loss, and Structural Similarity Index (SSIM). We also employ an image generation metric named Inception Score (IS) \cite{salimans2016improved}, which measures both the visual quality and generation diversity. Joining StackGAN \cite{zhang2017stackgan}, we employ inception models fine-tuned on the CUB and Oxford-102 datasets to calculate IS, which helps obtain a more accurate conditional label distribution for generated images. We mainly adopt the center-square mask because we want to mainly cover what the sentences describe ,\ie, the target objects, which mostly lie in the middle of images in our datasets. We also conduct free-form completion experiment on the CUB dataset, following the mask setting from \cite{yu2018free}. Overall, the results (see Table \ref{tab:quantitative}) verify the visual authenticity, global consistency of our results as well as the completion variety. We attribute these substantial improvements to the multi-grained reasoning blocks and progressive generation process.

 For the evaluation of whether the generated area is consistent with the lexical semantic context (text) and visual semantic context (unmasked area), we employ the initial pre-trained REG in the \textit{Supervision} stage to generate referring expressions and compare them with ground truth texts using standard Natural Language Generation metrics, BLEU \cite{papineni2002bleu}, METEOR \cite{banerjee2005meteor} , ROUGE\_L \cite{lin2004rouge} and CIDEr \cite{vedantam2015cider}. This idea is inspired by \cite{hong2018inferring}. Our model completes 128*128 holes within a 256*256 resolution image, while others complete 64*64 holes within a 128*128 resolution image, which is their default generation resolution. Overall, the results across various evaluation metrics consistently indicate that our proposed method achieves better results against three SOTA SIM models. The results on Referring Expression Generation (see Table \ref{tab:eval_caption}) explicitly show that our model can generate grounded completion faithful to text input.

\begin{figure}[!t] \begin{center}
    \includegraphics[width=1.0\columnwidth]{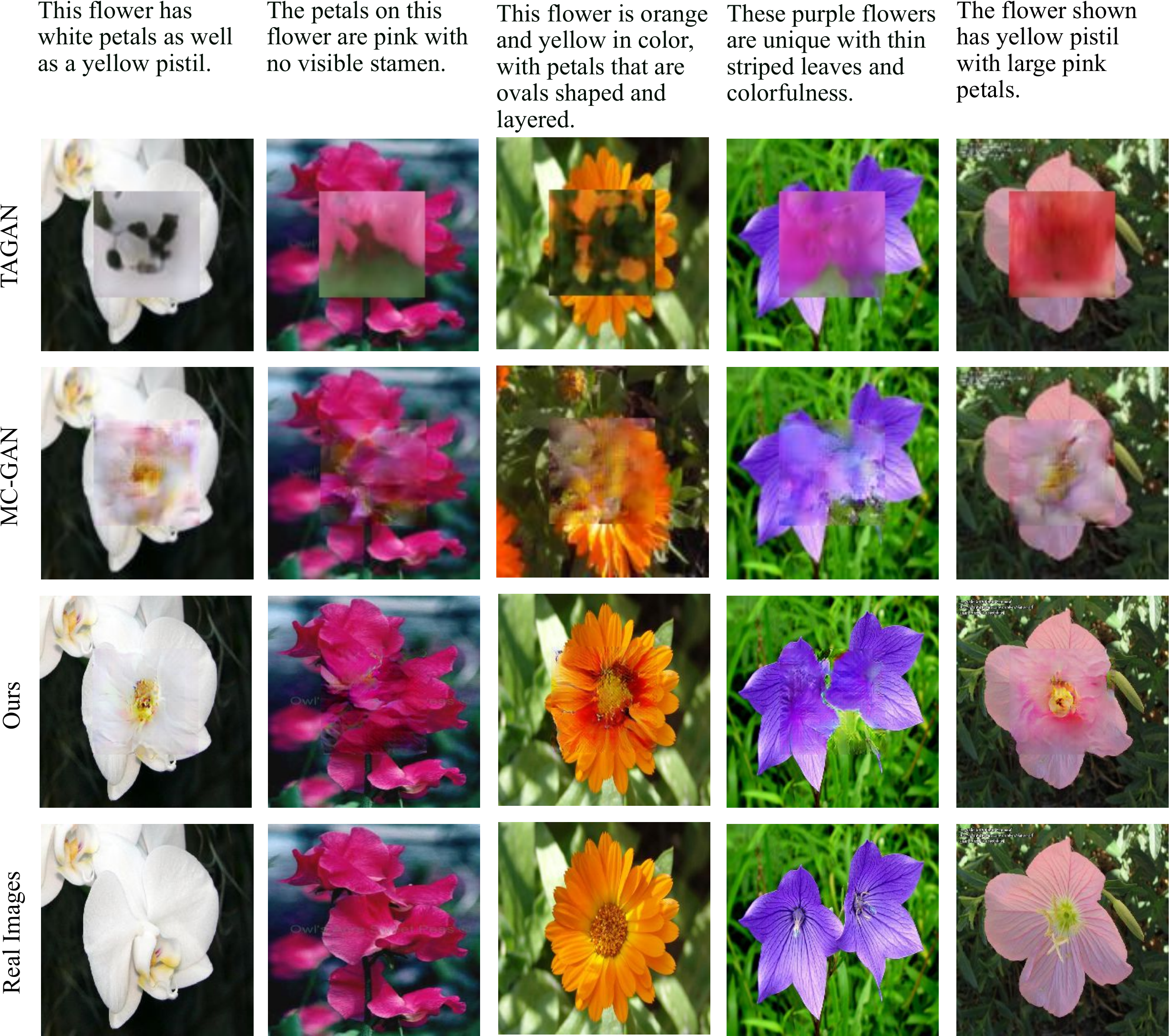}
    \caption{
    Qualitative comparison of three methods on the Oxford-102 test set.
	}
\label{fig:compareFlower}
\end{center} \end{figure}

	
\begin{figure}[!t] \begin{center}
    \includegraphics[width=1.0\columnwidth]{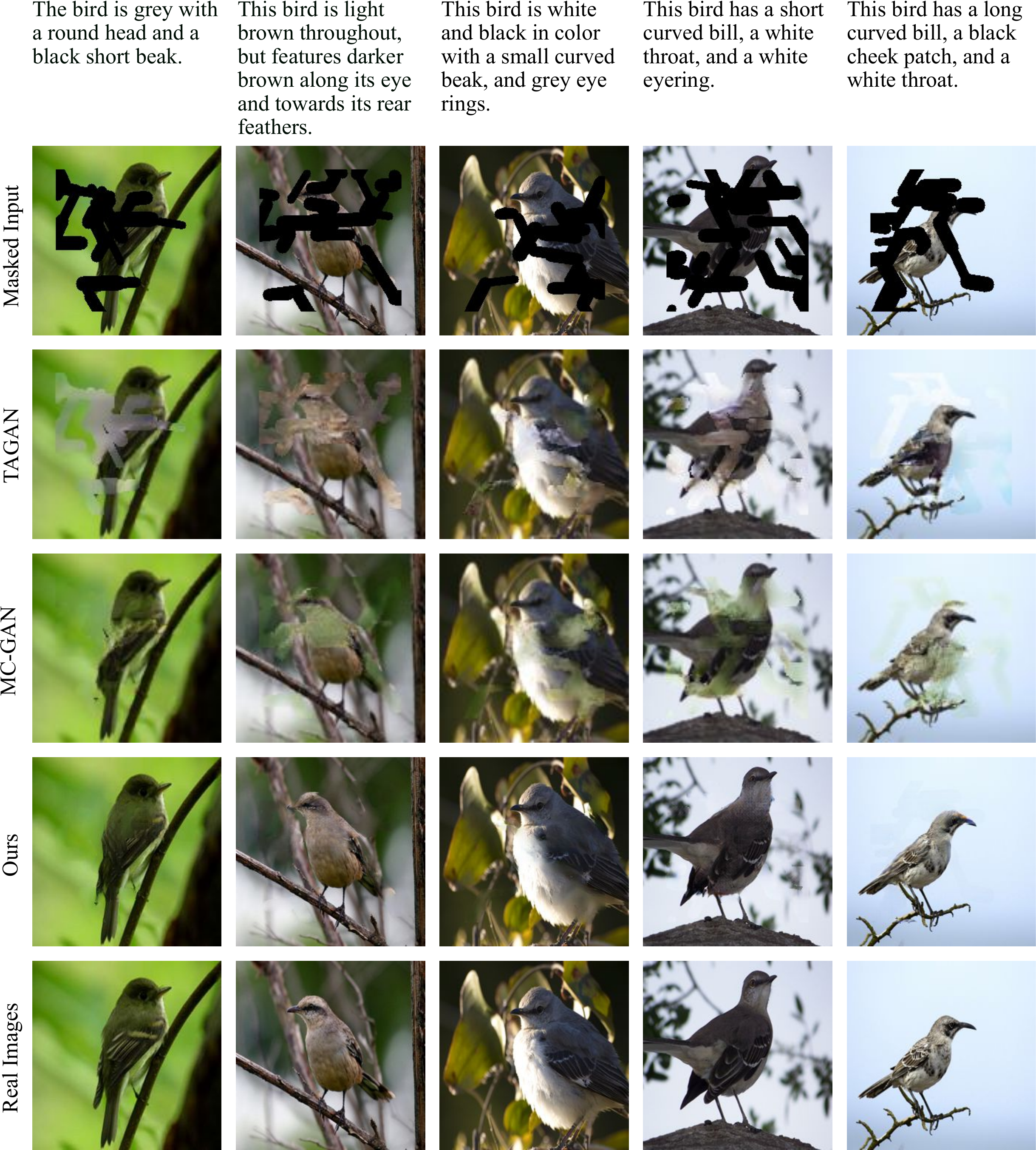}
    \caption{
    Qualitative results on the CUB test set with irregular mask setting.
	}
\label{fig:compareIrregular}
\end{center} \end{figure}

\subsection{Qualitative Evaluation}




\begin{figure}[!t]
\centering
\includegraphics[width=\columnwidth]{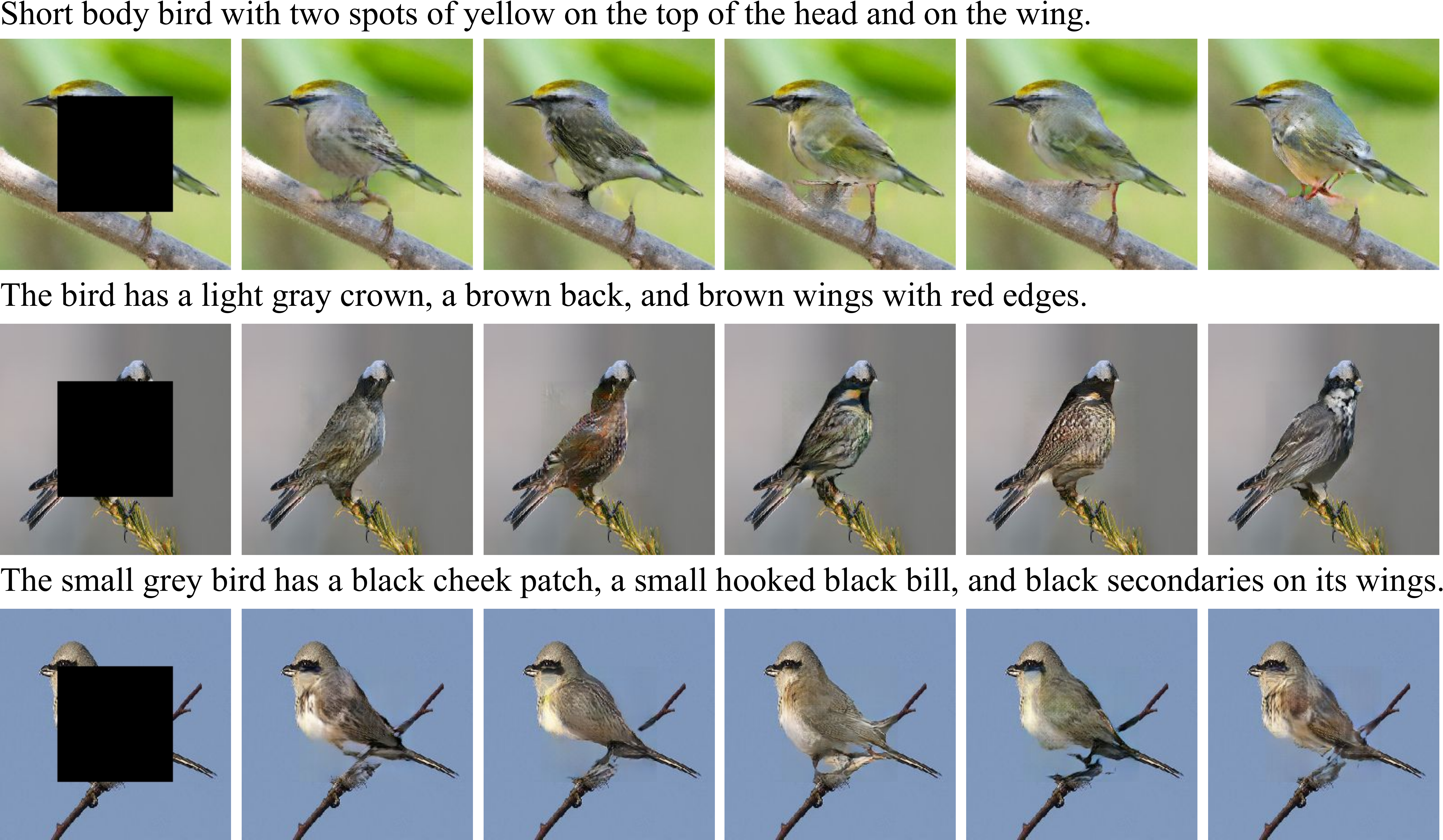}
\caption{
Qualitative evaluation of generation diversity.
}
\label{fig:variance}
\end{figure}

\subsubsection{Subjective Analysis}

Figure \ref{fig:compare} and Figure \ref{fig:compareFlower} displays the completion results produced by our proposed method and three modified comparison models concerning quality assessment. These samples are conditioned on text descriptions and center-masked images on the test dataset. Figure \ref{fig:compareFlower} further shows the free-form completion results on the CUB test set. Our method produces images with a coherent structure and vivid grounded details (\ie, factual attributes) in most cases, comparing to the Dong \etal, MC-GAN, and TAGAN.

Semantic image manipulation methods, which are often designed for changing a subtle attribute of the original image, may fail to generate consistent partial foreground and background structure, such as TAGAN. MC-GAN improves the performance by incorporating visual context (the background) into the generation process simultaneously. However, they both fail to effectively model the important lexical semantic context, resulting in the absence of many fine-grained details. Our model achieves better results in most cases, concerning both the visual authenticity, text-image consistency, and foreground-background continuity. It demonstrates the effectiveness of progressive reasoning modules for grounded completion.

\subsubsection{Generation Variety}

\begin{figure}[!t]
\centering
\includegraphics[width=\columnwidth]{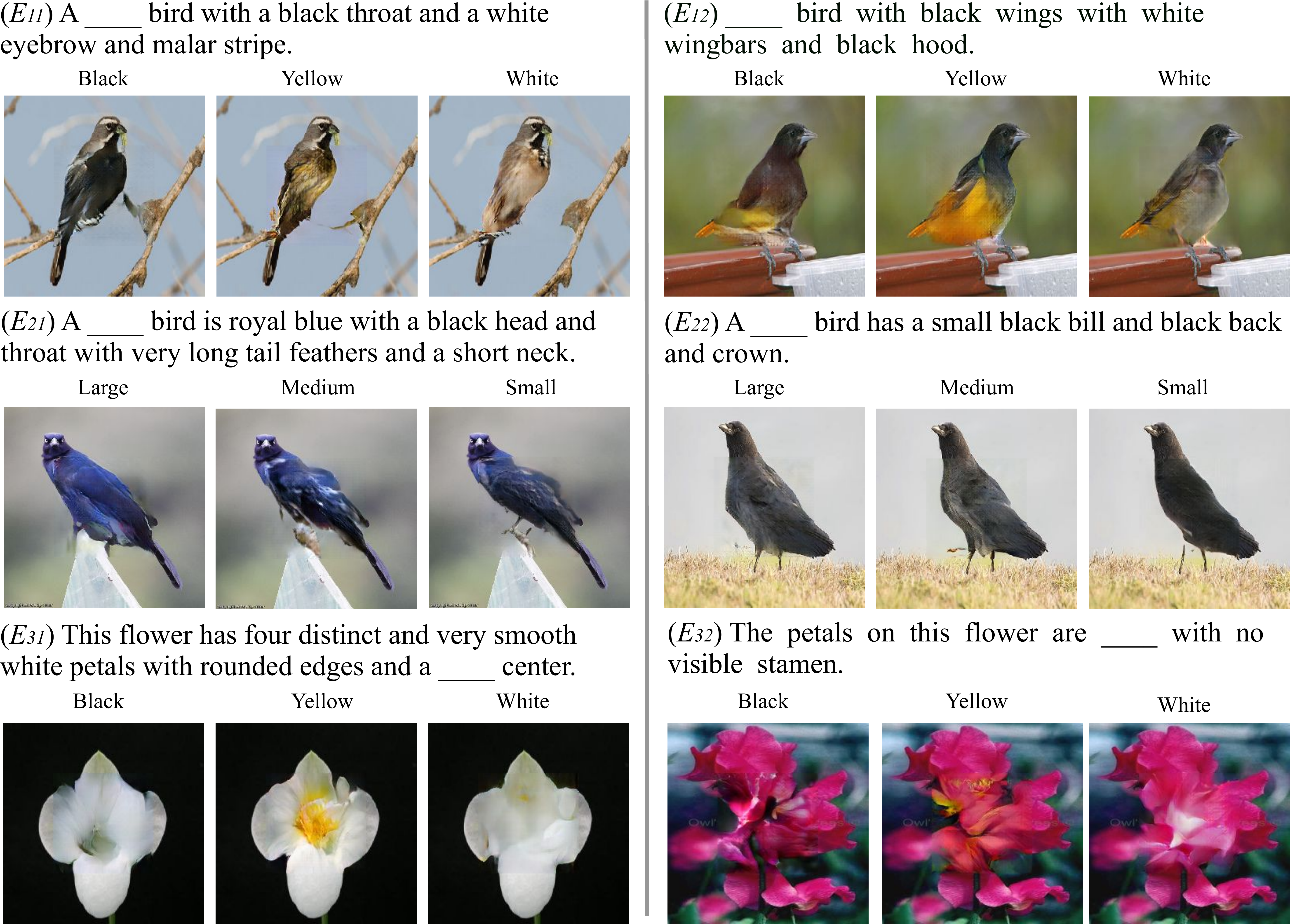}
\caption{
Controllable completion results by deliberately altering the attributes.
}
\label{fig:colorChange}
\end{figure}

Figure \ref{fig:colorChange} shows the controllable completion results. We deliberately change the factual attribute (\eg colors and sizes) within the input text. The results show that our model is able to capture the fine-grained semantic concepts and generate completions with corresponding details.

In order to evaluate the generation variety, we generate multiple samples given the same text-background inputs (see Fig. \ref{fig:variance}). While keeping faithful to the textual description and preserving visual continuity, our model generates diverse images with different texture, patterns, and color saturation. These results indicate that the proposed Creation path is a promising direction for better leveraging limited annotations and increase diversity in an unsupervised way.

\begin{table}[!t]
\centering
\caption{
    Quantitative evaluation results of referring expression generation on CUB dataset.
}
\setlength\doublerulesep{0.5pt}
\small
\begin{tabular}{l|cccc}
& \multicolumn{4}{c}{Referring Expression Generation}  \\
\multicolumn{1}{c|}{Method}        & BLEU-4         & METEOR   &ROUGE\_L          & CIDEr          \\
\hline \hline
Dong \etal     & 0.087   & 0.193   & 0.364   & 0.465 \\
\hline
TAGAN         & 0.096          & 0.196          & 0.368           & 0.513          
\\
\hline
MC-GAN          & 0.099          & 0.198          & 0.373           & 0.553          \\
\hline
Ours    & \textbf{0.102} & \textbf{0.199} & \textbf{0.376}  & \textbf{0.584} \\
\hline
Real Image    & 0.120    & 0.208    & 0.392     & 0.733
           \\
\hline
\end{tabular}
\label{tab:eval_caption}
\end{table}


\begin{table}[!t]
\centering
\caption{
Ablation test of different architectures.
}
\setlength{\tabcolsep}{2.5pt}
\setlength\doublerulesep{0.5pt}
\begin{tabular}{l|cccc}


\multicolumn{1}{c|}{Models} &   PSNR $\uparrow$    &    TV loss $\downarrow$ &  SSIM $\uparrow$   & IS  $\uparrow$      \\
\hline \hline

B(aseline)      & 19.20    &   12.53   &    0.801    &    3.71                 \\
 +R(easoning blocks)     & 19.31   &    11.84   &    0.810    &    3.84  \\
 +C(reation path)     & 19.41   &  11.89    &    0.813   &    4.09 
        \\
\hline
 +D(ual learning)    &  \textbf{19.68}    &   \textbf{10.73}   &    \textbf{0.819}   & \textbf{4.34}   \\
\hline
1 FGR   &  19.66    &  11.24   &   0.817   & 4.21   \\
0 FGR   & 19.41    &  11.54   &  0.811   & 3.86   \\
\hline

\end{tabular}
\label{table:ablation}
\end{table}

\section{Ablation Study}

To obtain a better understanding of different modules in our model, we surgically remove some components and construct different architectures (see Table \ref{table:ablation}). B denotes the baseline method, which only takes the masked image as input. R stands for the group of reasoning blocks, which includes CGR and FGR. C is the creation path without referring expression generator, which is named as D, \ie Dual Learning. The results indicate that the elimination of any component would result in a decrease in efficiency. To investigate whether the hierarchically stacked FGR blocks is beneficial, we gradually replace the last FGR block in our model with a plain resnet block, which takes only the visual features from the previous reasoning block as the input, \ie, without considering the semantic concepts. Our model includes two FGR. Therefore, 1 FGR indicates that the last FGR is replaced, and 0 FGR indicates that all FGRs are replaced by plain resnet blocks. The results verify the merit of our hierarchical architecture.

\begin{figure}[!t]
\centering
\includegraphics[width=\columnwidth]{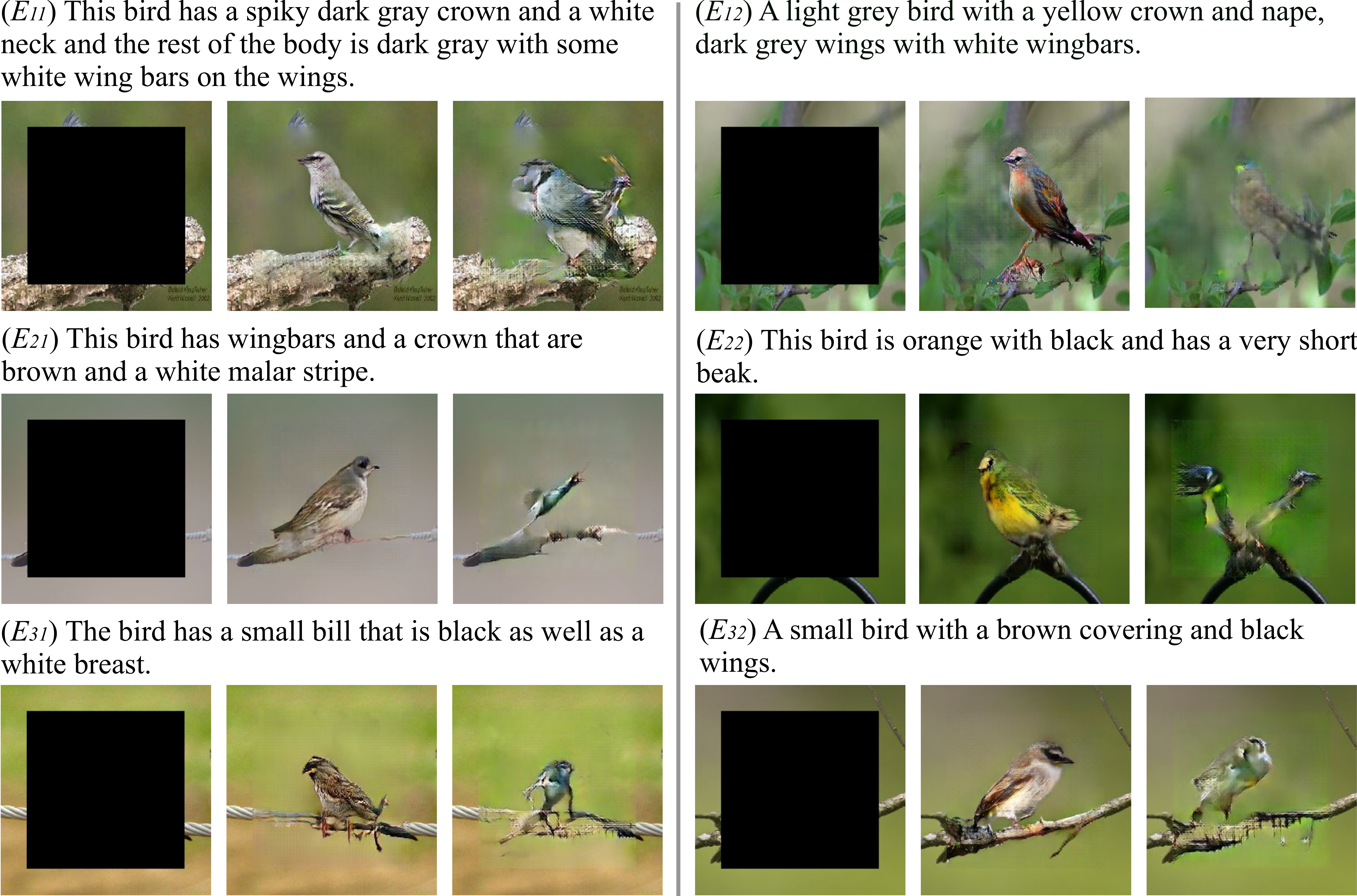}
\caption{
Completion results with less visual information (bigger holes). Masked input (Left), ours (Middle), baseline (Right). The baseline is constructed by removing text-relevant components.
}
\label{fig:biggerhole}
\end{figure}

By altering the center mask size form 50\% * 50\% to 75\% * 75\%, we also qualitatively evaluate the generation ability between our model and the baseline model (without text input) when given less visual semantic information (see Fig. \ref{fig:biggerhole}). With the auxiliary semantic information, our model can generate meaningful content with structure continuity. For example, in the results of the baseline method, the salient object (the bird) is missing, and other stuff (like the tree branch) is partially missing. Our model generates the entire bird while completing the background stuff. In most cases, our model depicts the fine-grained details such as the orange and body in $E_{22}$ and white wing bars in $E_{11}$. The result indicates the merit of proposed CGR and FGR.

\section{Conclusion}
	
In this paper, we propose a novel progressive reasoning network for the challenging Lexical Semantic Image Completion task, which aims to generate \textit{grounded} results faithful to the textual description and \textit{controllable} results by changing the attributes within the text. Our architecture encapsulates the \textit{coarse-grained reasoning block} and the \textit{fine-grained reasoning block} to model the structure of different modalities and perform structure alignment and translation. Besides conventional \textit{paired-reconstruction} generation, we incorporate the idea of Dual Learning and devise an \textit{unpaired-creation} path to mitigate the \textit{unimodal biases} problem and increase generalization. The consistent quantitative improvement across various metrics and substantial qualitative results on two fine-grained datasets reveal the efficacy of our proposed method.

\ifCLASSOPTIONcaptionsoff
  \newpage
\fi



%
%
%

\bibliographystyle{IEEEtran}
\bibliography{egbib}

%








\end{document}